\documentclass[10pt,twocolumn,letterpaper]{article}

\usepackage{cvpr}
\usepackage{times}
\usepackage{epsfig}
\usepackage{graphicx}
\usepackage{amsmath}
\usepackage{amssymb}
\usepackage{float}
\usepackage{tablefootnote}
\usepackage{multirow}
\usepackage{comment}
\usepackage{enumitem}
\usepackage{enumerate}
\usepackage{MnSymbol}
\usepackage{hyperref}
\usepackage{setspace,lipsum}

\begin{document}

\title{First Competition on Presentation Attack Detection on ID Card}

\author{ 
Juan E. Tapia\textsuperscript{$\dagger$}\textsuperscript{1},
Naser Damer\textsuperscript{$\dagger$}\textsuperscript{2,3},
Christoph Busch\textsuperscript{$\dagger$}\textsuperscript{1},
Juan M. Espin\textsuperscript{$\dagger$}\textsuperscript{4}, 
Javier Barrachina\textsuperscript{$\dagger$}\textsuperscript{4}, \\
Alvaro S. Rocamora \textsuperscript{$\dagger$}\textsuperscript{4}, 
Krištof Ocvirk \textsuperscript{5,*}, 
Leon Alessio \textsuperscript{5,*}, 
Borut Batagelj\textsuperscript{5,*}, 
Sushrut Patwardhan\textsuperscript{6,*},\\ 
Raghavendra Ramachandra\textsuperscript{*,6}, 
Raghavendra Mudgalgundurao \textsuperscript{*,6}, 
Kiran Raja\textsuperscript{*,6},\\ 
Daniel Schulz \textsuperscript{*,7}, 
Carlos Aravena \textsuperscript{*,7} \\ 
\textsuperscript{1}Hochschule Darmstadt (h\_da), da/sec-Biometrics and Internet Security Research, Germany.\\
\textsuperscript{2}Fraunhofer Institute for Computer Graphics Research IGD, Darmstadt, Germany,\\
\textsuperscript{3}Department of Computer Science, TU Darmstadt, Darmstadt, Germany,\\
\textsuperscript{4}Facephi company, Spain,\\
\textsuperscript{5} Faculty of Computer and Information Science, University of Ljubljana, Ljubljana, Slovenia,\\
\textsuperscript{6} Norwegian University of Science and Technology (NTNU), Gjøvik, Norway,\\
\textsuperscript{7} ID VisionCenter (IDVC), Santiago, Chile\\
\textsuperscript{$\dagger$}{\tt\small Organizer}
\textsuperscript{*}{\tt\small Competitors},
}


\maketitle
\thispagestyle{empty}

\begin{abstract}
This paper summarises the Competition on Presentation Attack Detection on ID Cards (PAD-IDCard) held at the 2024 International Joint Conference on Biometrics (IJCB 2024). The competition attracted a total of ten registered teams, both from academia and industry. In the end, the participating teams submitted five valid submissions, with eight models to be evaluated by the organisers. The competition presented an independent assessment of current state-of-the-art algorithms. Today, no independent evaluation on cross-dataset is available; therefore, this work determined the state-of-the-art on ID cards. To reach this goal, a sequestered test set and baseline algorithms were used to evaluate and compare all the proposals. The sequestered test dataset contains ID cards from four different countries. In summary, a team that chose to be "Anonymous" reached the best average ranking results of 74.80\%, followed very closely by the "IDVC" team with 77.65\%.  
\end{abstract}

\section{Introduction}

The accelerated evolution in consumer smartphone cameras and the COVID-19 pandemic has increased the interest in remote biometric verification systems. The capacity to reach the customer remotely for services such as e-commerce, digital banking, and general fintech requires robust systems for remote identity verification. One approach for this verification is using a picture of an official identity document, such as a national ID card, and comparing the data with a frontal face photograph (selfie) of the person in question, both captured remotely by the user "in the wild" condition. 

Remote identity verification processes can encounter attacks in which a user's identity is impersonated. Today, the simplest and most common attacks are presentation attacks. In a presentation attack, the attacker impersonates some of the captured samples, official documents, or selfies. With the intention to mitigate said problem, face presentation attack detection (face PAD) is a widely studied field. However, document attack detection is a new field and difficult to access due to the privacy of identity documents.

Nowadays, presentation attacks on ID cards, such as printing a photo or displaying it on a screen, are widespread. Also, the number of images available for training and testing is limited due to privacy concerns. As a result, many solutions trained on small datasets overfit intra-dataset conditions because the train and test set stem from the same source, limiting the generalisation capabilities.

PAD-IDCard 2024 is the first competition in the ID card series. It offers \textbf{(a)} an independent assessment of current state-of-the-art ID Card Presentation Attack Detection algorithms and \textbf{(b)} an evaluation protocol, including real printed-out and screen replay attacks and bona fide ID card images. Researchers can follow the evaluation protocol after the competition is closed to benchmark their solutions with PAD-IDCard winners and baselines. Today, there exists no independent evaluation where the approaches are evaluated in a cross-dataset scenario; hence, it is still being determined which methods, if any, perform well under such more realistic test conditions.

The rest of the article is organised as follows: Section~\ref{sec:related} summarises the related works of PAD in ID cards. Section~\ref{sec:databases} describes the datasets and depicts examples of images. Sections~\ref{sec:submission} and \ref{sec:eval} describe the submission and evaluation process. The metrics used for the evaluation and the proposed methods are described in Sections ~\ref{sec:eval} and ~\ref{sec:method}, respectively. The experiments and results are presented in Section~\ref{sec:exp-result}. Lastly, Section~\ref{sec:conclu} summarises the findings of the competition and discusses potential future work.

\section{Related work}
\label{sec:related}

In recent years, several works have proposed PAD methods to detect presentation attacks where ID cards are used to circumvent the security of remote verification systems~\cite{shi2019docface+, Tapia-eusipco}.

Berenguel et al. \cite{berenguel2017counterfeit} developed an application to classify ID documents forged by a scan-printing operation. Their application allows the capture of Spanish ID documents using a mobile device and the assessment of their validity. The counterfeit detection module apply texture descriptors, principal component analysis, and feature pooling to classify regions of interest using linear Support Vector Machines (SVM). The final decision of labelling a document as genuine or counterfeit is performed by a Naïve Bayes classifier. 


Gonzalez et al. \cite{gonzalez2021hybrid} presented a two-stage method for detecting tampered ID cards, which was trained and evaluated on a dataset with real Chilean national ID cards. The proposed method uses a pre-trained MobileNet model\cite{howard2017mobilenets} to detect borderlines in the photo ID zone caused by composite tampering, while a second lightweight CNN, termed ``BasicNet'', was trained from scratch to detect the physical source of the document. 

Mudgalgundurao et al. \cite{mudgalgundurao2022pixelwise} proposed to adapt a pixel-wise supervision model in \cite{DBLP:conf/isvc/DamerSFBKK21} that is used, along with a binary classification objective, to train presentation attack detectors on an in-house dataset of German ID cards and residence permits. The proposed system uses a simplified DenseNet \cite{huang2017densely} architecture, which the authors compare against baseline face PAD approaches.

Chen et al. \cite{chen2022domain} employed a scheme based on Siamese networks for document recapture detection. The network is trained on triplets of patches extracted from bona fide, recaptured, reference documents. A custom ``forensics loss'' aimed at attracting genuine and reference representations while repelling recaptured and reference representations. The authenticity of a questioned document is evaluated using the distance metrics from three triplets. The authors created a synthetic university student ID card dataset to test their system.

Benalcazar and Tapia et al. \cite{Benalcazar} explored the effectiveness of computer vision algorithms and generative models for data augmentation while training fraud detection networks. The authors propose populating templates with synthetic data to create additional bona fide presentations and training a StyleGAN-ADA network to generate synthetic bona fide samples from scratch.
 
Magge et al. \cite{magee2023investigation} explored the application of the Meijering filter \cite{meijering2004design} for detecting recaptured identity documents. The authors created a dataset of recaptured images based on the publicly available BID \cite{alysson2020bid} dataset and used it to train an SVM classifier on the raw histogram data obtained using the filter. Although their system does not compare well with approaches that utilise neural networks, it remains an attractive alternative due to being transparent and explainable.

Most of the aforementioned studies trained and tested their proposed systems on private datasets using bona fide presentations of ID cards obtained from Government entities, company services, and banks. As such, it is difficult to scrutinise and improve upon these systems since the data can not be distributed publicly due to privacy concerns. 

Currently, open-set datasets like MIDV 500 \cite{MIDV-500}, MIDV 2019 \cite{midv-2019}, MIDV 2020 \cite{MIDV2020AC} and DLC 2021 \cite{dlc2021}, despite offering a rich amount of country representations and document types, fall short due to their limited number of unique user identities and few examples of bona fide and screen displays attacks on ID cards. Conversely, private datasets are not available to compare the results. These fundamental limitations undermine the potential of PAD ID card models to accurately learn and generalise across the wide variability inherent in ID cards. 

As a starting point, the quantity of unique user data is crucial for teaching models to discern between bona fide and fake documents. It is a complicated task because of the minor subject base of these datasets. Table~\ref{tab:survey-table} summarises the most relevant datasets in this field and can be used as a starting point to train a PAD system.

\begin{table}[H]
\centering
\caption{Summary ID card datasets available in the State of the art.}
\label{tab:survey-table}
\resizebox{\columnwidth}{!}{%
\begin{tabular}{|c|c|c|c|c|}
\hline
Author            & Datasets   & Images  & User   & Comments                                                                 \\ \hline
Soares et al. \cite{alysson2020bid}     & BID-Data    & 28,800   & 8      & \begin{tabular}[c]{@{}c@{}}Synthetic data\\ No Genuine data\end{tabular} \\ \hline
Mudgalgundurao et al. \cite{mudgalgundurao2022pixelwise}.       & Private     & 104,882 & 86     & Genuine ID card                                                          \\ \hline
González et al.\cite{gonzalez2021hybrid}   & Private     & 54,980  & 5,000  & Genuine ID card                                                          \\ \hline
González et al.\cite{gonzalez2023improving}   & Private     & 190,000 & 16,000 & Genuine ID card                                                          \\ \hline
Benalcazar et al.\cite{Benalcazar} & Private     & 38,477  & 9,286  & Genuine ID card                                                          \\ \hline
Markham et al. \cite{markham2023openset}   & Open-set & 500     & 50     & Generated from templates - Transfer style                                \\ \hline
\begin{tabular}[c]{@{}c@{}}Arzalov et al. \cite{MIDV-500}\\ Bulatov et al.\end{tabular} &
  \begin{tabular}[c]{@{}c@{}}Open-set\\ MIDV-500 \cite{MIDV-500}\\ MIDV-2020 \cite{MIDV2020AC}\end{tabular} &
  \begin{tabular}[c]{@{}c@{}}500\\ Videos\end{tabular} &
  50 &
  No Genuine ID card \\ \hline
Polevoy et al. \cite{dlc2021} &
  \begin{tabular}[c]{@{}c@{}}Open-set\\ DLC2021\end{tabular} &
  1000 &
  1000 &
  \begin{tabular}[c]{@{}c@{}}No Genuine ID card\\ Generated from templates\end{tabular} \\ \hline
Koliaskina et al. \cite{midv-holo} &
  \begin{tabular}[c]{@{}c@{}}Open-set\\ MID Holo\end{tabular} &
  \begin{tabular}[c]{@{}c@{}}700\\ Video\end{tabular} &
   &
  \begin{tabular}[c]{@{}c@{}}No Genuine ID card - Utopia ID card\\ 300 holographic - 400 videos\end{tabular} \\ \hline
Park et al. \cite{Park}&
  \begin{tabular}[c]{@{}c@{}}Open-Set\\ KID-2K\end{tabular} &
  34,662 &
  82 &
  \begin{tabular}[c]{@{}c@{}}No Genuine ID card\\ For 46 people who do not exist\\ **\end{tabular} \\ \hline
\end{tabular}%
}
\end{table}


\begin{figure*}[]
\centering
\includegraphics[scale=0.09]{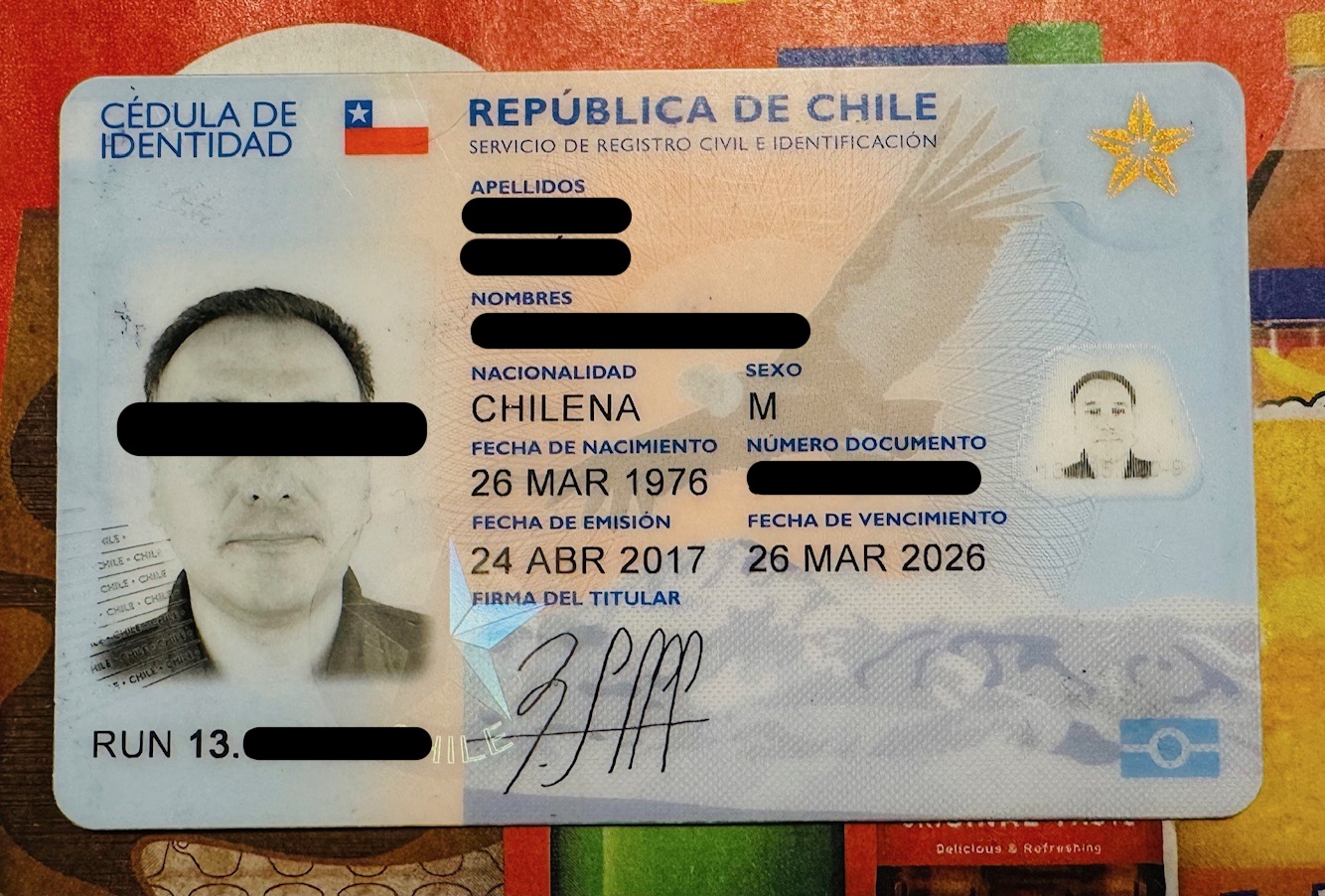} 
\includegraphics[scale=0.083]{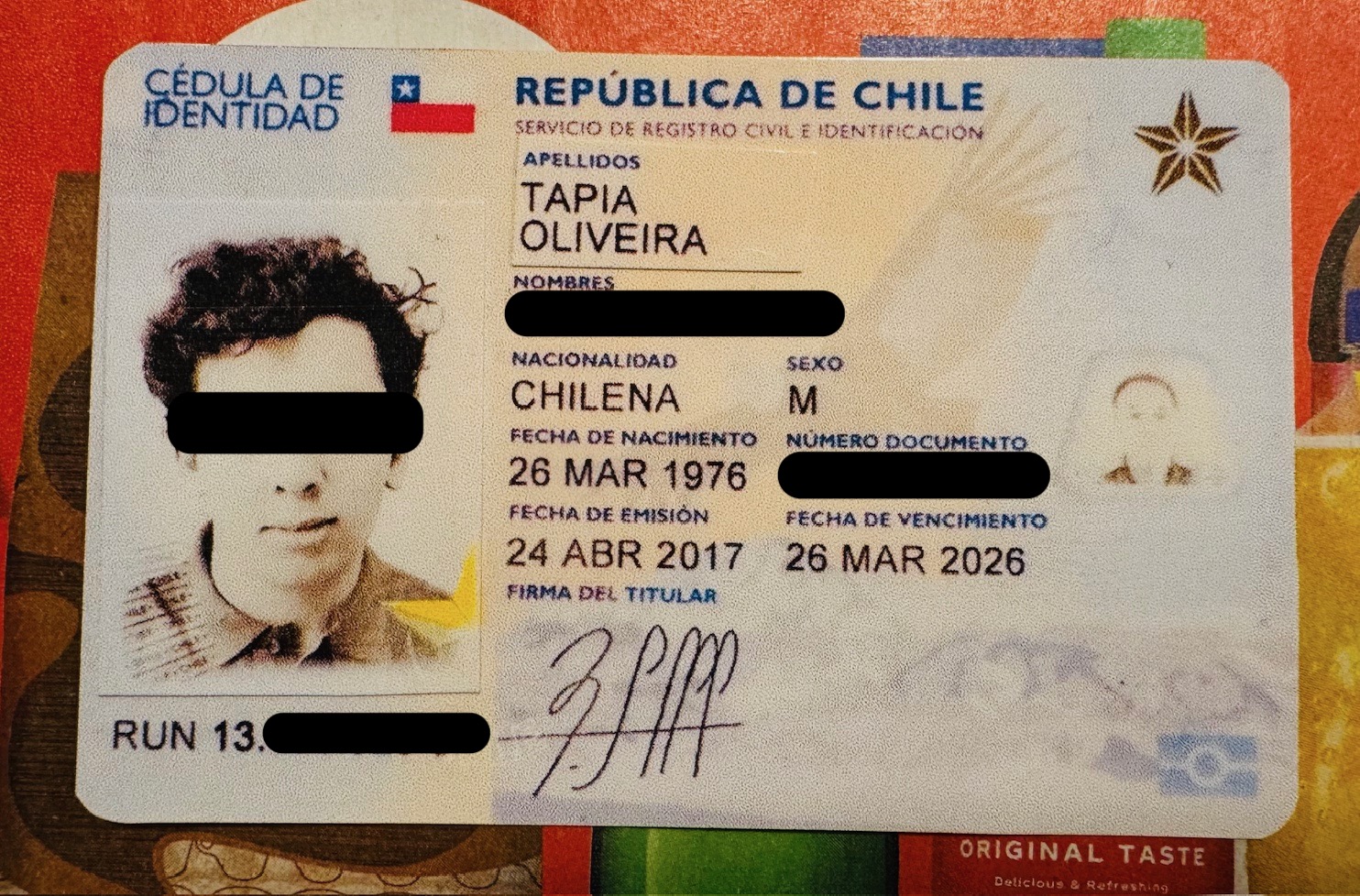} 
\includegraphics[scale=0.09]{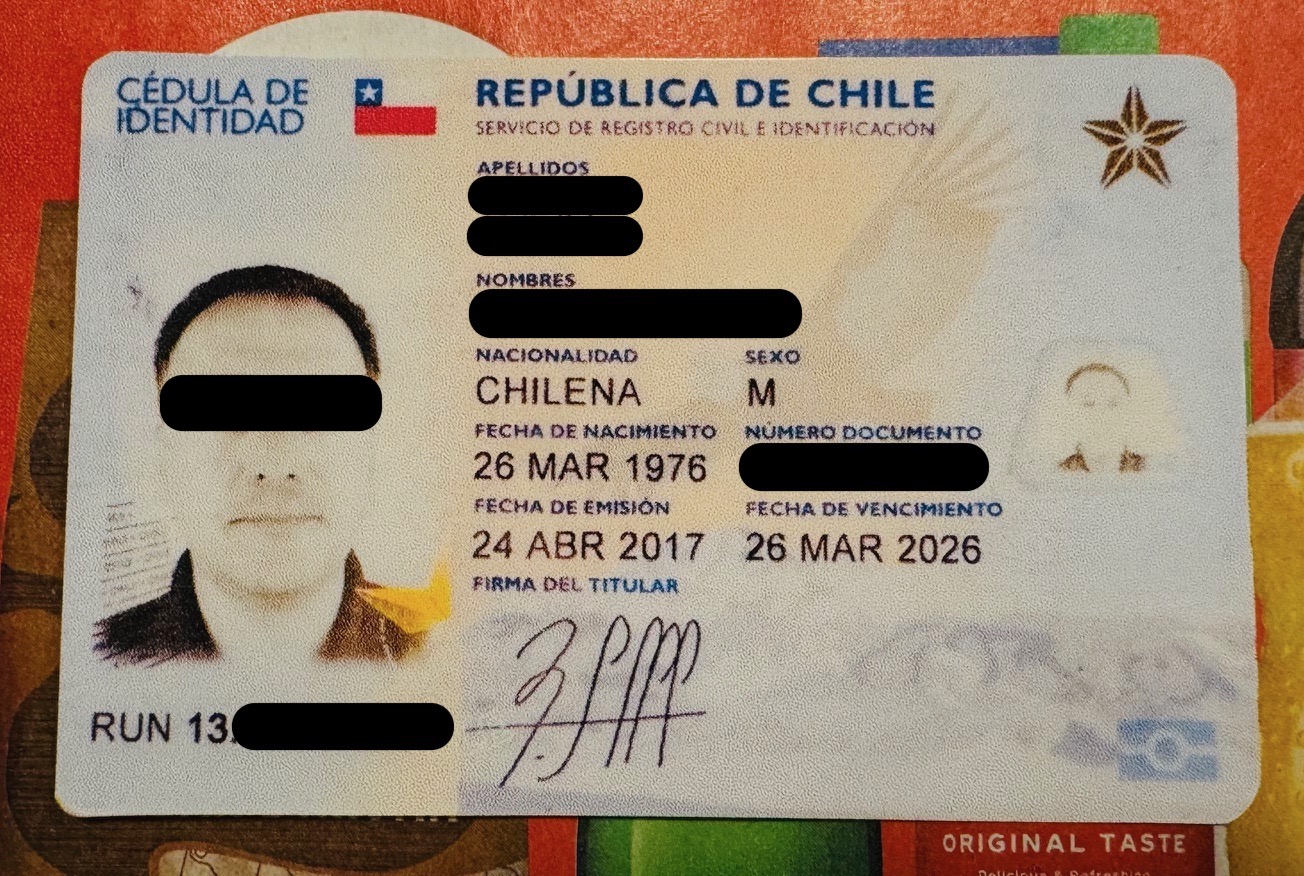}
\includegraphics[scale=0.08]{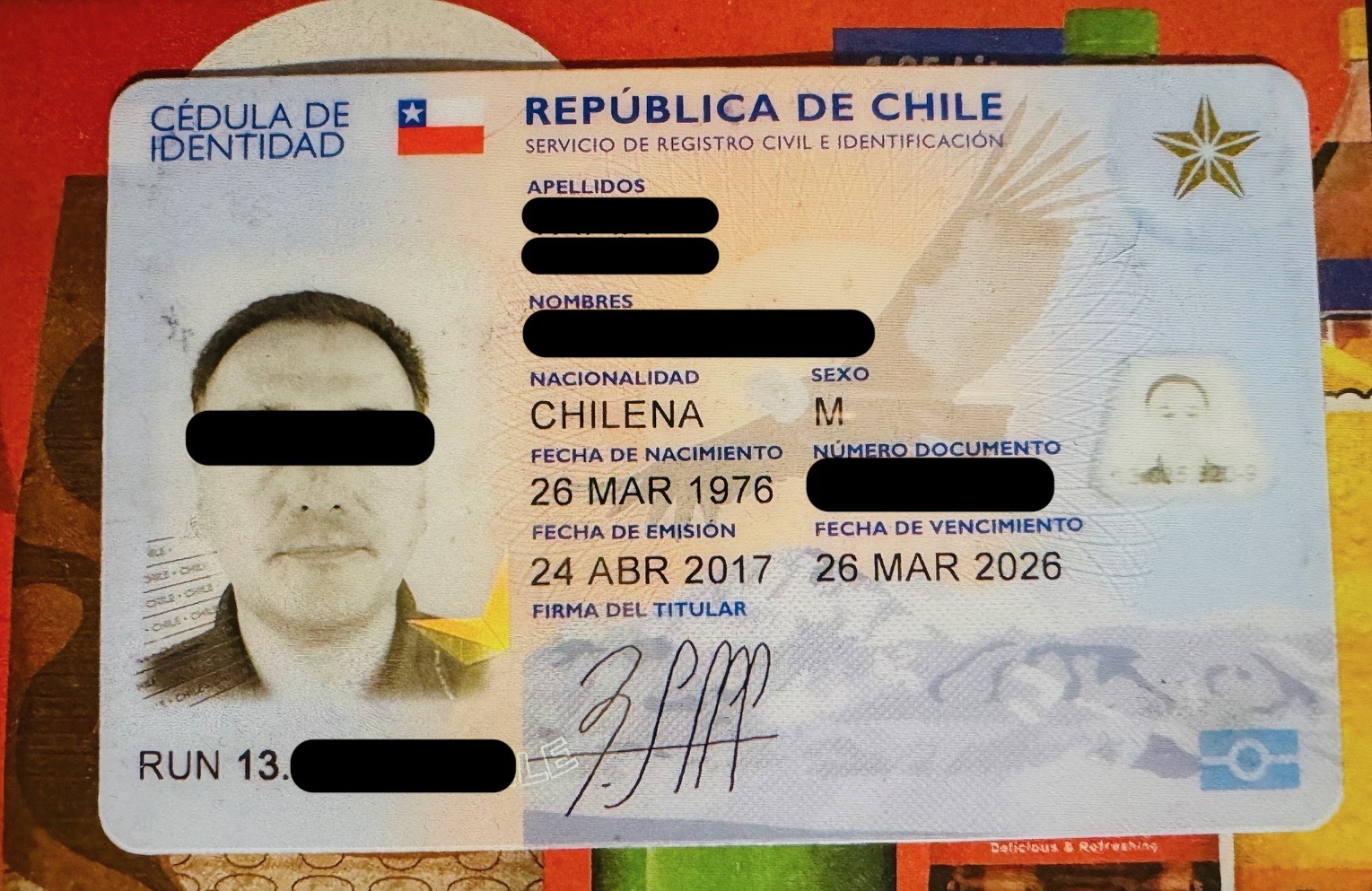} 
\caption{Example of images used to validate the ID card PAD model. Left to right: Bona fide, Composite, Printed and Screen.}
\label{fig:examples}
\end{figure*}


Given the significant limitations of available public datasets for studying PAD card applications, a private dataset for training and testing was created for this competition from digital users' documents to develop a baseline evaluation. The dataset was generated in-house by h\_da and the Spanish company Facephi.


\section{Datasets}
\label{sec:databases}

No training datasets were provided to the participants for the competition. Each team used any available training dataset, such as open-set, private, or synthetic datasets, for research and commercial purposes. 

For the development of the baseline method, we created a private training dataset consisting of ID cards from four countries: Spain (ESP), Chile (CHL), Argentina (ARG), and Costa Rica (CRI). This set contains bona fide images, which represent pictures captured directly from a genuine ID card. The composite attack represents an ID card image modified by swapping the face or the text area between two ID cards. The composite attacks were created manually and automatically based on data augmentation techniques. The print attacks were created using an ID card printed out on glossy paper and PVC cards with different resolutions. The screen attacks were made by capturing ID card displays on various screens such as tablets, smartphones, and laptops.

The test set partition was sequestered for all the participants, and it consisted of ID cards from four different countries: Chile (CHL), Guatemala (GUA), Panama (PAN), and Mexico (MEX). For each country bona fide and attack images are contained. All the attacks were created manually and automatically, following the same conditions as those previously defined for the private training dataset used for the baseline methods. These images also present different qualities, which means visual artefacts in the area of face photos and high-quality images without any visual artefacts. The printed attacks were created from glossy paper and PVC cards. The screen images were captured from several sources and resolutions, such as tablets, smartphones, and laptop screens.
Some of the ID card images are ICAO compliant, and others are not. This competition did not test injection attacks. 

One small set of 4 images was provided to all the teams that submitted a model in order to validate the results. This set contains one ID card image of each type: bona fide, composite, print, and screen ID card image, as shown in Figure \ref{fig:examples}.

Tables \ref{tab:db-baseline-exp1}, \ref{tab:db-baseline-exp2}, and \ref{tab:db-baseline-exp3} describe the details of each group of images used for the experiments 1, 2, and 3 explained in Section \ref{sec:method}.

\section{Submission process}
\label{sec:submission}

All the participants submitted a link (e.g., a link to Google Drive or Dropbox) with a folder with the compiled model in a “conda environment” with the file “enviroment.yml” for Python 3.8, where it indicates all the libraries necessary to run their model. This model accepted as input a “CSV file” with the path input images as a “filename”. The output from the model was another CSV file with three columns, “filename”, “score”, and “class”, as Presentation Attack Detection (PAD) for all the “test” samples. The value 0 means “bona fide”, and the value 1 means “attacks”. The evaluation Python file provided should accept two parameters: The path to the evaluation\_list.csv (--evaluation) and the output path of the scores (--output).

\section{Evaluation Criteria}
\label{sec:eval}
The detection performance of biometric PAD algorithms is standardised by ISO/IEC 30107-3\footnote{\url{https://www.iso.org/standard/79520.html}}. The most relevant metrics for this study are Attack Presentation Classification Error Rate (APCER), Bona fide Presentation Classification Error Rate (BPCER), and BPCER\textsubscript{AP}. Those metrics determine the error rates when classifying an instance between bona fide and the different Presentation Attack Instrument Species (PAIS).  

The APCER metric measures the percentage of attack presentations incorrectly classified as bona fide for each different PAIS. The worst-case scenario is considered when evaluating an entire system. The computation method is detailed in Equation~\ref{eq:apcer}, where the value of $N_{PAIS}$ corresponds to the number of attack presentation images, $RES_{i}$ is $1$ if the $i$th image is classified as an attack, or $0$ if it was classified as a bona fide presentation according to a predefined threshold.

\begin{equation}\label{eq:apcer}
    {APCER_{PAIS}}=1 - \frac{1}{N_{PAIS}}\sum_{i=1}^{N_{PAIS}}RES_{i}
\end{equation}

On the other hand, the BPCER metric measures the proportion of bona fide presentations wrongly classified as attacks. The BPCER can be computed using Equation~\ref{eq:bpcer}, where $N_{BF}$ is the amount of bona fide presentation images, and $RES_{i}$ takes the same values described earlier for the APCER metric. The two metrics determine the system's performance and are subject to a specific operation point. 

\begin{equation}\label{eq:bpcer}
    BPCER=\frac{\sum_{i=1}^{N_{BF}}RES_{i}}{N_{BF}}
\end{equation} 

Finally, BPCER\textsubscript{AP} and the Equal Error Rate (EER) are used to analyse the PAD system's performance for a specific operating point. The latter is the operating point where APCER and BPCER are equal. This operating point corresponds to the intersection with the diagonal line in a Detection Error Trade-off (DET) curve, which is also reported for all the experiments. On the other hand, the BPCER\textsubscript{AP} is the BPCER value when the APCER is $100/AP$. In this work, we use: BPCER\textsubscript{10}, BPCER\textsubscript{20} and BPCER\textsubscript{100}, which correspond to APCER values of 10\%, 5\% and 1\%, respectively.

An average ranking determines the winning team. A weighting factor was selected to increase the metric's contribution in the most challenging operational points, such as BPCER\textsubscript{100}. The team with the lowest $AV_{Rank}$ won the competition. This metric weighted the BPCER\textsubscript{10,20,100} as follows:

\begin{equation}\label{eq:avrank}
\scriptstyle
    AV_{rank}=BPCER_{10}\times0.2+ BPCER_{20}\times0.3 + BPCER_{100}\times0.5
\end{equation} 

\section{Methods}
\label{sec:method}

\subsection{Baseline description}
\label{sec:baseline}

Three baselines were defined, according to the available datasets, to explore the real conditions and possible scenarios of today's state of the art:

\begin{description}
    \item[Baseline 1:] Training with a private dataset (bona fide and attack) and evaluating with the sequestered test dataset.  

    \item[Baseline 2:] Training with only bona fide from private datasets plus only attacks from open-set datasets (MIDV-500 and MIDV-Holo) and evaluating with the sequestered test dataset.

    \item[Baseline 3:] Training with a mixture of private and open-set datasets such as MIDV-500 and MID-Holo, evaluating with the sequestered test dataset. 
\end{description}

The preprocessing pipeline for the images input into our networks begins with the segmentation and alignment of the ID card present in each input image, following the method proposed in \cite{markham2023openset}. This initial step ensures that the ID card is correctly isolated and positioned, providing a consistent starting point for further processing. The images are reshaped to $384\times384$ for both experiments. Once the ID card is properly segmented, aligned, and reshaped, a series of augmentations are applied to the dataset to improve the robustness of the models. These augmentations include horizontal flipping to simulate variations in orientation and adjustments in brightness and contrast to account for different lighting conditions. Additionally, random JPEG compression is applied to introduce variations in image quality, mimicking real-world scenarios where image compression artefacts may be present. Using a random Gaussian filter helps simulating different levels of blur, while random hue adjustments and grey-scale conversions further diversify the dataset by altering the colour properties of the images. These preprocessing steps collectively ensure that the models are trained on a wide variety of image conditions, enhancing their ability to generalise and perform accurately across different scenarios.

For this competition, three different architectures were evaluated for each baseline experiment:  

MobileVITv2\footnote{\url{https://github.com/leondgarse/keras_cv_attention_models/tree/main}}, Efficientv2-S\footnote{\url{https://github.com/google/automl/tree/master/efficientnetv2}}, and MobileNetv3-Large\footnote{\url{https://github.com/kuan-wang/pytorch-mobilenet-v3}}. 
In all of them, the ImageNet weights were used to initialise and train all the models. All the baseline models were trained with three classes: bona fide, composite, print plus PVC, and screen. The bona fide images (64,933) are common in the three baselines. The MobileVITv2 reached the best performance and is used to compare with the competitors in the rest of the work. 
  
A sequestered dataset with a total of 21,000 images was created for evaluation with ID cards from four countries and four PAIs. Table \ref{tab:db-baseline-exp1} provides an overview of the dataset used for Baseline 1, which considers only private datasets for training.

\begin{table}[H]
\scriptsize
\centering
\caption{Baseline 1 dataset - Training/Validation using a private dataset}
\label{tab:db-baseline-exp1}
\resizebox{\columnwidth}{!}{%
\begin{tabular}{|c|c|c|c|c|}
\hline
          & Train & Val & Test & Total \\ \hline
Bona fide &   64,933    &  11,458  &   5,000  &  81,391     \\ \hline
Composite &   90,101    &  15.900  &   5,000  &  111,001     \\ \hline
Print     &   30,398    &  5,364   &   6,000  &  41,762     \\ \hline
Screen    &   82,232    &  14,511  &   5,000  &  101,743     \\ \hline
Total     &   267,664   &  47,233  &   21,000 &  335,897     \\ \hline
\end{tabular}%
}
\end{table}


\begin{table}[H]
\scriptsize
\centering
\caption{Baseline 2 dataset - Training/Validation on bona fide presentations from a private dataset and attacks from public datasets}
\label{tab:db-baseline-exp2}
\resizebox{\columnwidth}{!}{%
\begin{tabular}{|c|c|c|c|c|}
\hline
          & Train & Val & Test & Total \\ \hline
Bona fide &  64,933   &  11,458 &  5,000  &  81,391     \\ \hline
Composite &  9,283    &  1,638  &  5,000  &  15,921     \\ \hline
Print     &  26,623   &  4,698  &  6,000  &  37,321     \\ \hline
Screen    &  19,608   &  3,460  &  5,000  &  28,068     \\ \hline
Total     &  120,447  &  21,254  & 21,000 &  162,701     \\ \hline
\end{tabular}%
}
\end{table}

Table \ref{tab:db-baseline-exp2} provides an overview of the dataset used for Baseline 2, which considers only open-set datasets for training plus private bona fide presentations. This experiment considers MIDV-500 and MIDV-Holo.

Table \ref{tab:db-baseline-exp3} provides an overview of the dataset used for Baseline 3, which considers private datasets (267,664 images) plus open-set datasets for training (55,514 images). This experiment considers MIDV-500 and MIDV-Holo.

\begin{table}[H]
\scriptsize
\centering
\caption{Baseline 3 dataset - Training/Validation on a mixture of private and public datasets}
\label{tab:db-baseline-exp3}
\resizebox{\columnwidth}{!}{%
\begin{tabular}{|c|c|c|c|c|}
\hline
          & Train & Val & Test & Total \\ \hline
Bona fide &   64,933   &  11,458   &   5,000  &  81,391     \\ \hline
Composite &   99,384   &  17,538   &   5,000  &  121,922     \\ \hline
Print     &   57,021   &  10,062   &   6,000  &  73,083     \\ \hline
Screen    &   101,840  &  17,971   &   5,000  &  124,811     \\ \hline
Total     &   323,178  &  57,029   &   21,000 &  401,207     \\ \hline
\end{tabular}%
}
\end{table}

\subsection{Submission and Team proposals}
Ten teams were registered for the competition. However, five different teams have submitted their models for evaluation. In total, 11 models were evaluated, 8 submitted by the teams, plus three baselines. Each team described its own proposed method.

\textbf{Team Asmodeus} This team from NTNU proposed two models, AsmodeusV1 and AsmodeusV2, based on Dynamic Snake convolution (DSC) \cite{Qi_2023_ICCV} for detecting high-frequency artefacts in added-in print photos, screen photos, and during GAN synthesis.
DSC uses a deformed kernel to learn high-frequency noise. The model consists of sequential DSC modules with a fully connected layer at the end for classification.

\textbf{Team "Anonymous"} This team that chose to be Anonymous, proposed two models for the Document Presentation Attack Detection (DocPAD) based on the MobilenetV3-large Convolutional Neural Network (CNN) \cite{mbv3}. This CNN was trained to detect artefacts commonly found in display and print attacks. The model was trained using academic datasets such as MIDV, DLC, and internal data developed by their QA team. The Anonymous team was leveraging its capabilities. The system aims to accurately identify and prevent presentation attacks, ensuring the integrity and reliability of identity verification processes.

\textbf{Team FRIFE} This team from the University of Ljubljana developed a model for detecting composite attacks. The team decided to develop a line detection algorithm based on traditional computer vision techniques with the goal of detecting visible lines in unusual places on the ID cards. For recapture attacks, they trained a single Xception-based model \cite{xception} to detect both screen and print-out recaptures.

\textbf{Team IDVC-PAD-IDCARD} For this challenge, the IDVisionCenter (IDVC) company team proposed models called IDVC\_V1 and IDVC\_V2. The two models are based on a pipeline composed of an ID-Card detection followed by a PAD algorithm. For the ID-Card detector, they trained a network based on the YOLO algorithm, using international open-set datasets and also their private dataset, using a resolution of $416\times416$ pixels. Then, for the PAD algorithm, they trained a network based on MobileNet\cite{mbv3}, using four distinct attack classes along with the bona fide class. The resolution for the PAD detection algorithm is $224\times224$ pixels. They manually created three presentation attack instruments: Printed, Replay (Screen), and Composite. They also created a presentation attack instrument automatically, consisting of swapping the face image for different ID cards. The difference between IDVC\_V1 and IDVC\_V2 is that V2 includes open-set datasets in the training stage. 

\textbf{Team Secure-ID} This second team from NTNU as well proposed a method based on the MIDV-500 dataset that served as the foundation for the competition. To generate synthetic ID cards, the GitHub repository \footnote{\url{https://github.com/Oriolrt/SIDTD_Dataset}} was utilised. The Canon TS-5000 printer was employed for print attacks, utilising papers of various qualities along with colour and grayscale images for detection purposes. Two Android smartphones and two monitors under different backgrounds and lighting settings were used to create replay attacks. A pixel-wise classification method is proposed to detect presentation attacks of the printed and digitally replayed attacks. The approach to using pixel-wise supervision is to leverage minute cues on various artefacts, such as Moiré patterns and artefacts left by the printers \cite{moire}.

\section{Experiment and Results}
\label{sec:exp-result}

This challenge involved two kinds of evaluations. The first evaluation addresses the three baseline methods, which were fine tuned on the private dataset after being pre-trained network as described in Section \ref{sec:baseline}. The second evaluation compares all submissions from all teams. 

The sequestered test set is common for all the submissions. Thus, each team trained its own model, but all were evaluated on the same test set. The test dataset is composed of ID cards from 4 different countries.

All the submissions were evaluated as a binary model to determine the winning team, which means bona fide versus attacks. As complementary information, the sequestered test dataset was evaluated separately by countries, i.e. Chile, Guatemala, Panama, and Mexico.

\begin{figure*}[]
    \centering
    \includegraphics[scale=0.35]{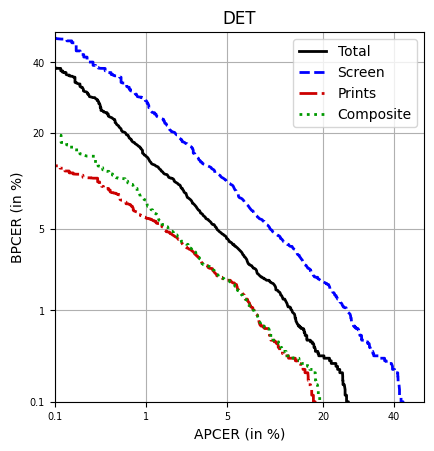} 
    \includegraphics[scale=0.35]{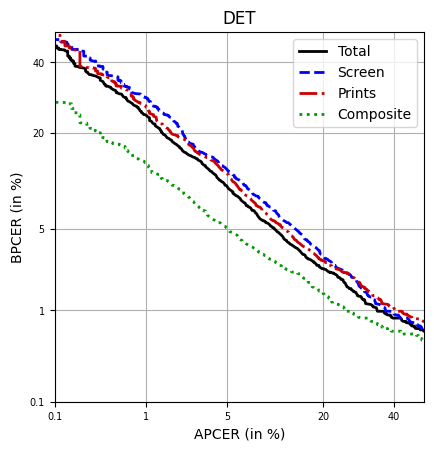} 
    \includegraphics[scale=0.35]{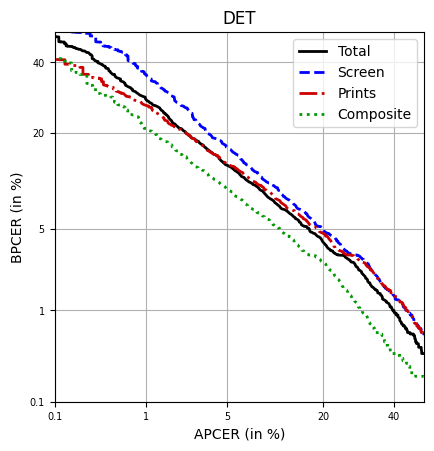} 
    \caption{DET Baseline results for MobileVIT models. The left to right plots show Baselines 1, 2, and 3 results, respectively. The black line represents the binary results of bona fide versus attacks.}
    \label{fig:baseline-for-3exp}
    \end{figure*}

\begin{figure*}[]
    \centering
    \includegraphics[scale=0.39]{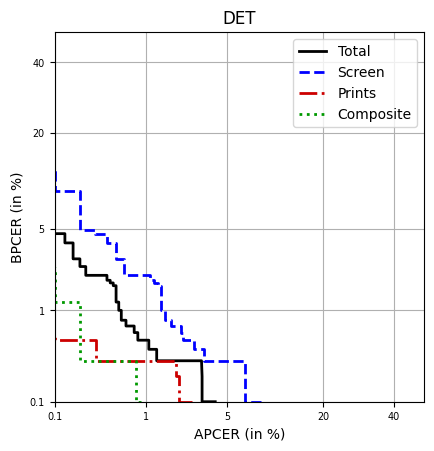}
    \includegraphics[scale=0.39]{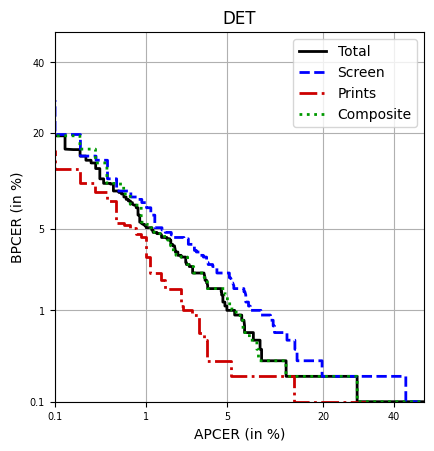}
    \includegraphics[scale=0.39]{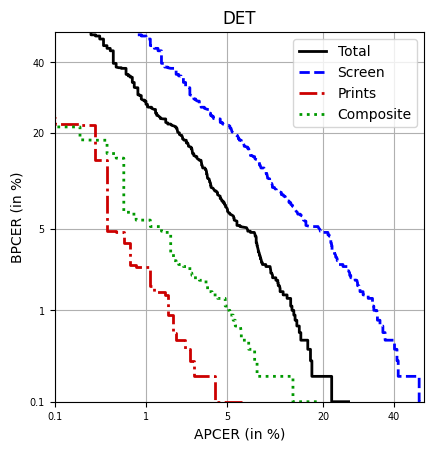}
    \includegraphics[scale=0.39]{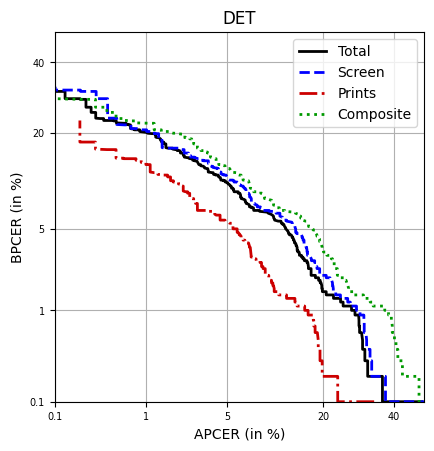}    
    \caption{DET Baseline results for MobileVIT models. The left to right plots show \textbf{Baseline 1} for 4 different ID card countries, Chile, Guatemala, Panama, and Mexico subsets. The black line represents the binary results of bona fide versus attacks.}
    \label{fig:base_exp1}
\end{figure*}

\begin{figure*}[!ht]
    \centering
    \includegraphics[scale=0.39]{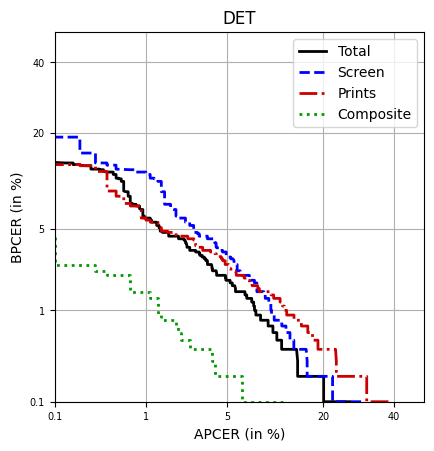}
    \includegraphics[scale=0.39]{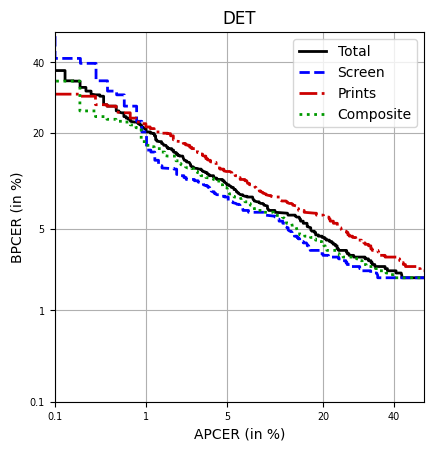}
    \includegraphics[scale=0.39]{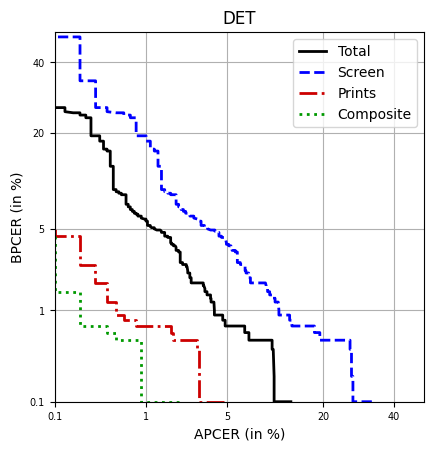}
    \includegraphics[scale=0.39]{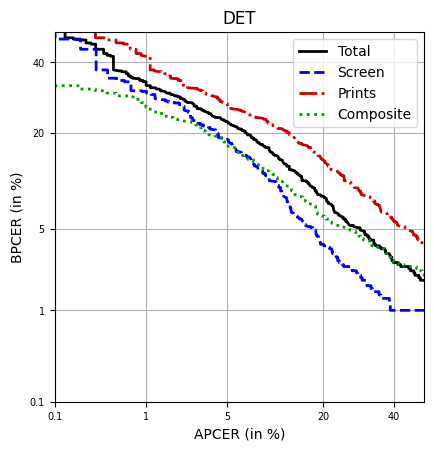}    
    \caption{DET Baseline results for MobileVIT models. The left to right plots show \textbf{Baseline 2} for 4 different ID card countries, Chile, Guatemala, Panama, and Mexico subsets. The black line represents the binary results of bona fide versus attacks.}
    \label{fig:base_exp2}
\end{figure*}

\begin{figure*}[!h]
    \centering
    \includegraphics[scale=0.38]{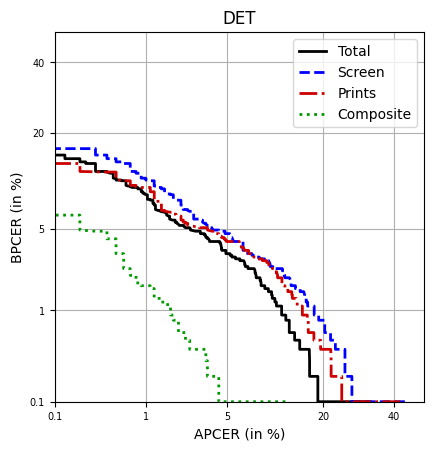}
    \includegraphics[scale=0.39]{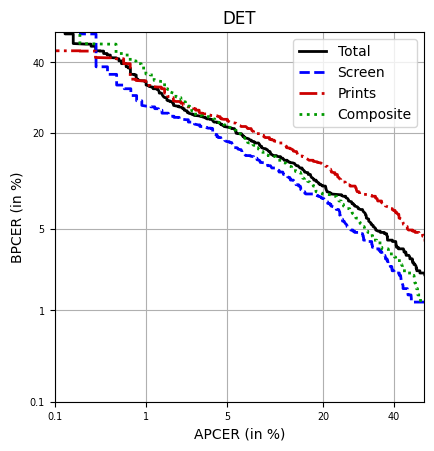}
    \includegraphics[scale=0.39]{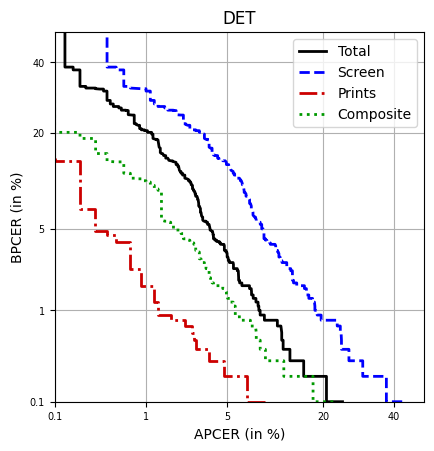}
    \includegraphics[scale=0.39]{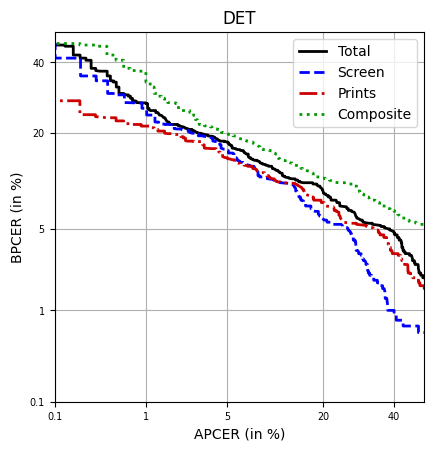}
    \caption{DET Baseline results for MobileVIT models. The left to right plots show \textbf{Baseline 3} for 4 different ID card countries, Guatemala, Panama, and Mexico subsets. The black line represents the binary results of bona fide versus attacks.}
    \label{fig:base_exp3}
\end{figure*}

Figure \ref{fig:baseline-for-3exp} shows the DET curves for the baselines 1, 2, and 3. The black line considers bona fide presentations versus all the attacks (i.e., composite, print, and screen). Each plot also includes the analysis of each PAIS isolated for research purposes. The green curve shows the composite attack, the red represents the printed attack, and the blue represents a screen attack. A single analysis by baselines 1, 2, and 3 are depicted in Figures \ref{fig:base_exp1}, \ref{fig:base_exp2} and \ref{fig:base_exp3}.

Further, the green curve shows that the composite attack reached a lower error rate for baseline 1, which means training in private datasets and testing in sequestered datasets. For baselines 2 and 3, the PAIS shows similar results for each one. 

\begin{figure*}[]
    \centering
    \includegraphics[scale=0.30]{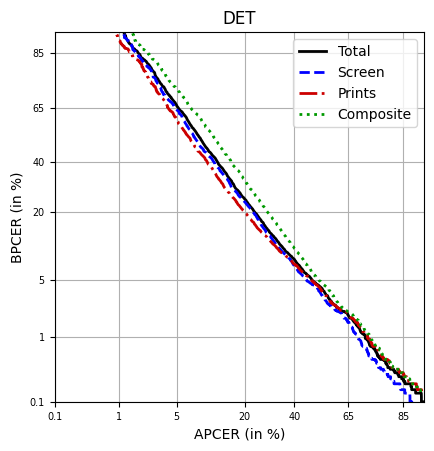}
    \includegraphics[scale=0.30]{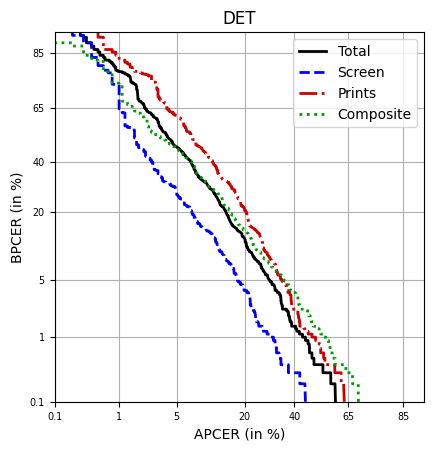}
    \includegraphics[scale=0.30]{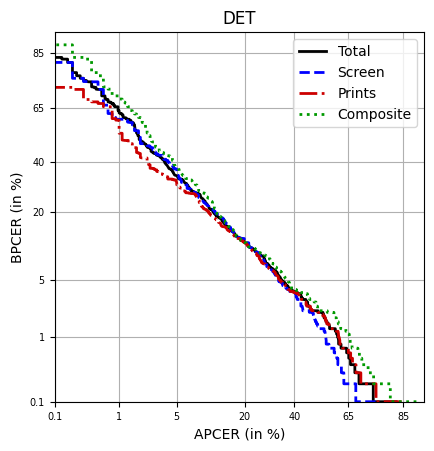}
    \includegraphics[scale=0.30]{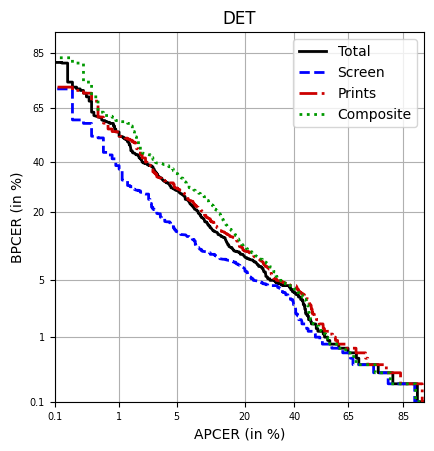}
    \includegraphics[scale=0.30]{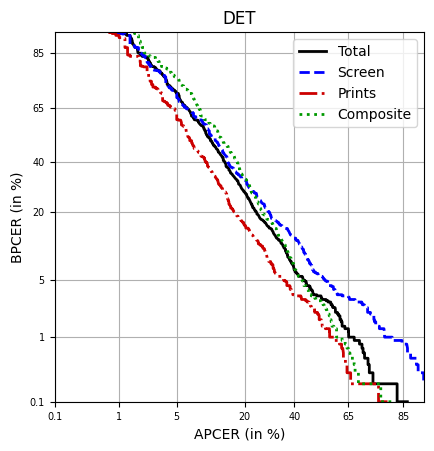}
    \caption{DET result for\textbf{ Anonymous\_V1} model for test set. The left-to-right plots show all countries together, followed by Chile, Guatemala, Panama, and Mexico subsets. The black line represents the binary results of bona fide versus attacks.}
    \label{fig:Anonimus}
    \end{figure*}

\begin{figure*}[!h]
    \centering
    \includegraphics[scale=0.30]{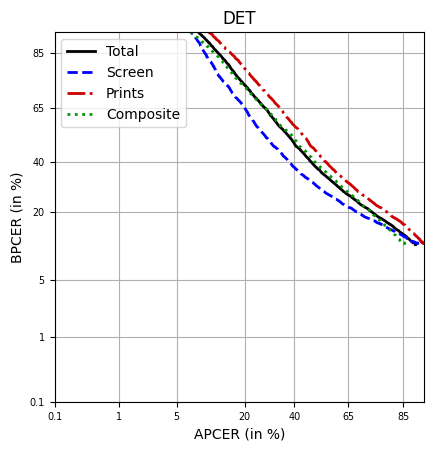}
    \includegraphics[scale=0.30]{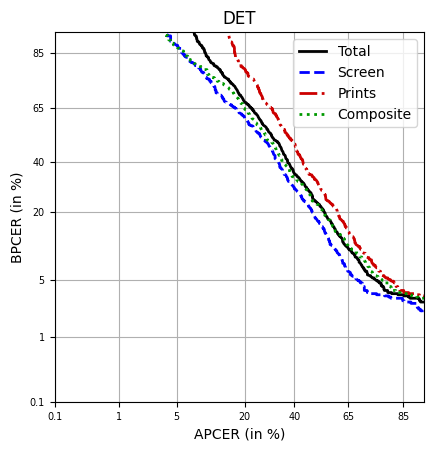}
    \includegraphics[scale=0.30]{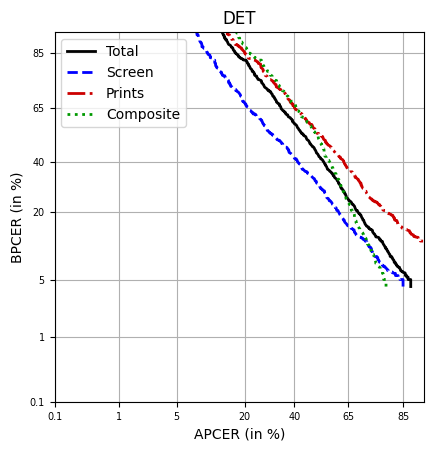}
    \includegraphics[scale=0.30]{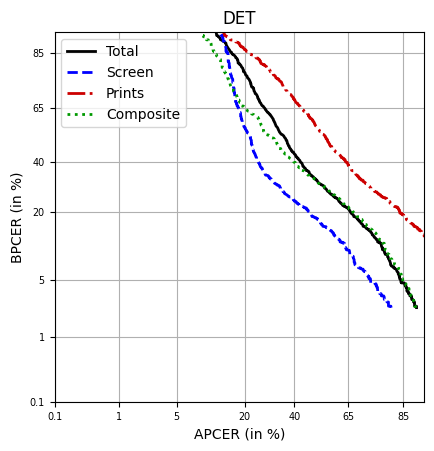}
    \includegraphics[scale=0.30]{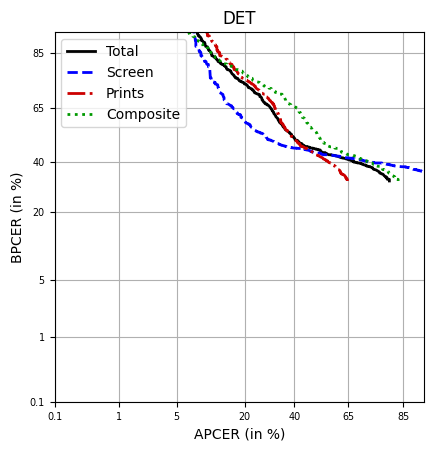}
   \caption{DET result for \textbf{FRIFE} model for the test set. The left-to-right plots show all countries together, followed by Chile, Guatemala, Panama, and Mexico subsets. The black line represents the binary results of bona fide versus attacks.}
    \label{fig:FRIFE}
    \end{figure*}

\begin{figure*}[!h]
    \centering
    \includegraphics[scale=0.30]{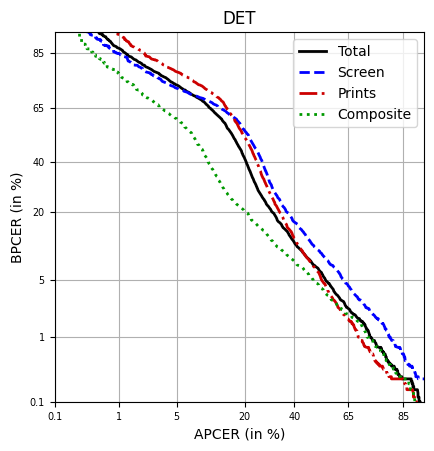}
    \includegraphics[scale=0.30]{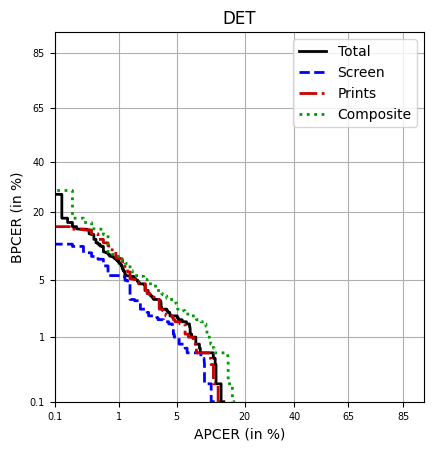}
    \includegraphics[scale=0.30]{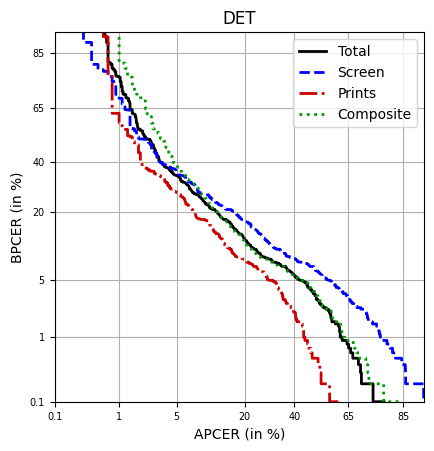}
    \includegraphics[scale=0.30]{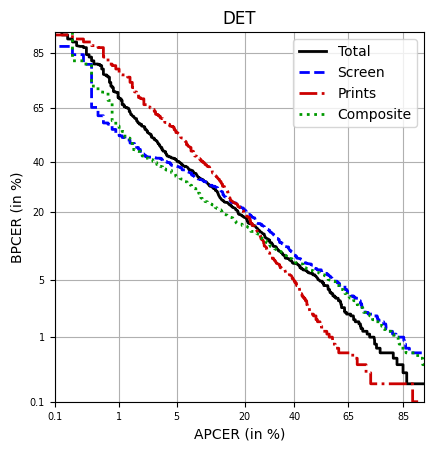}
    \includegraphics[scale=0.30]{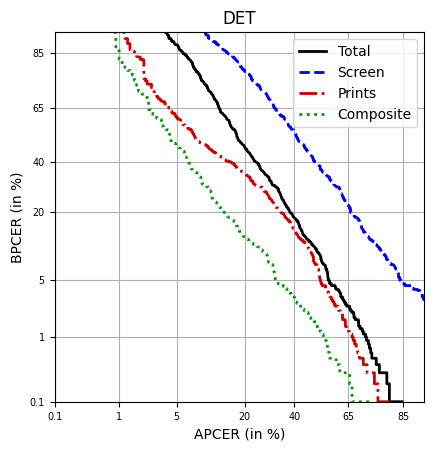}
    \caption{DET result for \textbf{IDVC\_V2} model for the test set. The left-to-right plots show all countries together, followed by Chile, Guatemala, Panama, and Mexico subsets. The black line represents the binary results of bona fide versus attacks.}
    \label{fig:IDVC}
    \end{figure*}

\begin{figure*}[!h]
    \centering
    \includegraphics[scale=0.29]{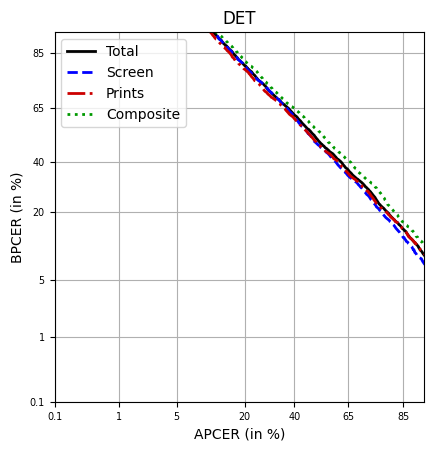}
    \includegraphics[scale=0.29]{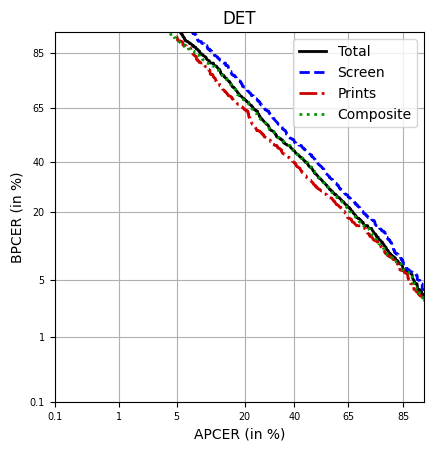}
    \includegraphics[scale=0.29]{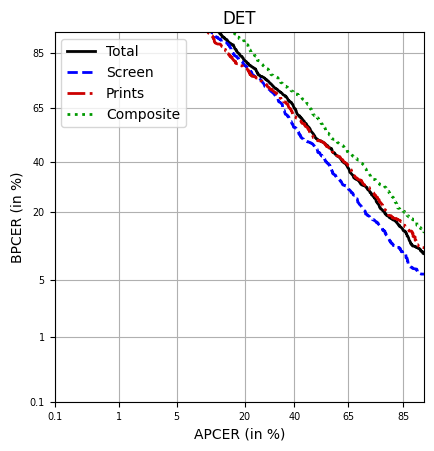}
    \includegraphics[scale=0.29]{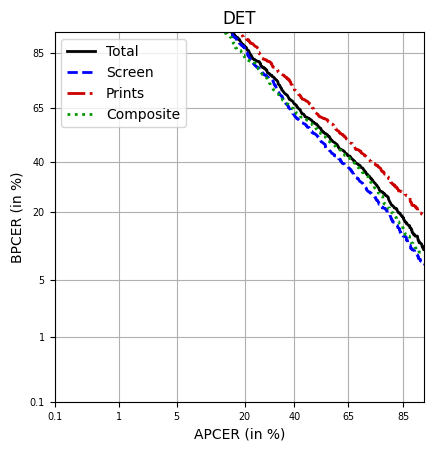}
    \includegraphics[scale=0.29]{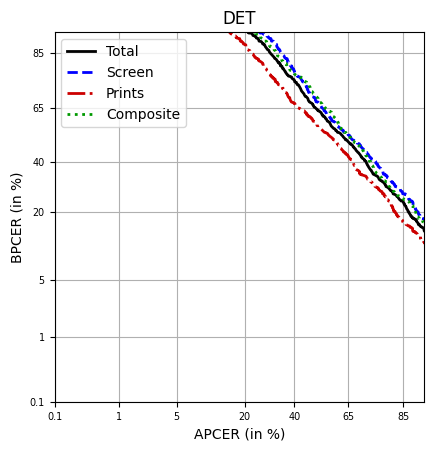}
    \caption{DET result for \textbf{SecureID} model for the test set. The left-to-right plots show all countries together, followed by Chile, Guatemala, Panama, and Mexico subsets. The black line represents the binary results of bona fide versus attacks.}
    \label{fig:Secure}
\end{figure*}

Figure \ref{fig:Anonimus}, \ref{fig:FRIFE}, \ref{fig:IDVC}, and \ref{fig:Secure} show the DET curve for the Anonymous, FRIFE, IDVC\_V2, and Secure-ID teams, respectively, for all countries evaluated together, followed by the single evaluation for Chile, Guatemala, Panama, and Mexico subsets.

In summary, the best results were reached by the Anonymous\_V1 team submission with an $AV_{Rank}$ of 74.30\% for the bona fide versus attack of all the countries together. The Asmodeus team reached the lowest results because both submitted models always reported the same score, i.e. 0.86 and 0.99 for Asmodeus\_V1 and Asmodeus\_V2, respectively. Both models presented by this team can not generalise well to different attacks reaching a higher $AV_{Rank}$.

Table \ref{tab:summary-results} shows all the submitted models' summary results evaluated based on Average Rank. The last column show the overall rank. 

\begin{table}[H]
\centering
\scriptsize
\caption{Summary submission results.}
\label{tab:summary-results}
\resizebox{\columnwidth}{!}{%
\begin{tabular}{ccccccc}
\hline
\multicolumn{1}{|c|}{\begin{tabular}[c]{@{}c@{}}Team\\ Name\end{tabular}} &
  \multicolumn{1}{c|}{\begin{tabular}[c]{@{}c@{}}EER\\ (\%)\end{tabular}} &
  \multicolumn{1}{c|}{\begin{tabular}[c]{@{}c@{}}BPCER10\\ (\%)\end{tabular}} &
  \multicolumn{1}{c|}{\begin{tabular}[c]{@{}c@{}}BPCER20\\ (\%)\end{tabular}} &
  \multicolumn{1}{c|}{\begin{tabular}[c]{@{}c@{}}BPCER100\\ (\%)\end{tabular}} &
  \multicolumn{1}{c|}{\begin{tabular}[c]{@{}c@{}}Average\\ Rank \\(eq. \ref{eq:avrank}) (\%)\end{tabular}} &
  \multicolumn{1}{c|}{Rank} \\ \hline
\multicolumn{7}{c}{\textbf{PAD-IDCard 2024 Competing Algorithms}} \\ \hline
\multicolumn{1}{|c|}{\textbf{Anonymous\_V1}} &
  \multicolumn{1}{c|}{21.87} &
  \multicolumn{1}{c|}{46.06} &
  \multicolumn{1}{c|}{65.82} &
  \multicolumn{1}{c|}{90.70} &
  \multicolumn{1}{c|}{\textbf{74.30}} &
  \multicolumn{1}{c|}{1} \\ \hline
\multicolumn{1}{|c|}{Anonymous\_V2} &
  \multicolumn{1}{c|}{29.01} &
  \multicolumn{1}{c|}{63.36} &
  \multicolumn{1}{c|}{76.82} &
  \multicolumn{1}{c|}{92.22} &
  \multicolumn{1}{c|}{81.82} &
  \multicolumn{1}{c|}{4} \\ \hline
\multicolumn{1}{|c|}{Asmodeus\_V1} &
  \multicolumn{1}{c|}{N/A\tablefootnote{The Asmodeus\_V1 model always delivers the score 0.86}} &
  \multicolumn{1}{c|}{N/A} &
  \multicolumn{1}{c|}{N/A} &
  \multicolumn{1}{c|}{N/A} &
  \multicolumn{1}{c|}{N/A} &
  \multicolumn{1}{c|}{7} \\ \hline
\multicolumn{1}{|c|}{Asmodeus\_V2} &
  \multicolumn{1}{c|}{N/A\tablefootnote{The Asmodeus\_V2 model always delivers the score 0.99}} &
  \multicolumn{1}{c|}{N/A} &
  \multicolumn{1}{c|}{N/A} &
  \multicolumn{1}{c|}{N/A} &
  \multicolumn{1}{c|}{N/A} &
  \multicolumn{1}{c|}{7} \\ \hline
\multicolumn{1}{|c|}{FRIFE} &
  \multicolumn{1}{c|}{44.09} &
  \multicolumn{1}{c|}{87.96} &
  \multicolumn{1}{c|}{93.06} &
  \multicolumn{1}{c|}{99.92} &
  \multicolumn{1}{c|}{95.47} &
  \multicolumn{1}{c|}{5} \\ \hline
\multicolumn{1}{|c|}{IDVC\_V1} &
  \multicolumn{1}{c|}{22.96} &
  \multicolumn{1}{c|}{65.40} &
  \multicolumn{1}{c|}{74.60} &
  \multicolumn{1}{c|}{84.38} &
  \multicolumn{1}{c|}{77.65} &
  \multicolumn{1}{c|}{2} \\ \hline
\multicolumn{1}{|c|}{IDVC\_V2} &
  \multicolumn{1}{c|}{25.91} &
  \multicolumn{1}{c|}{66.10} &
  \multicolumn{1}{c|}{74.42} &
  \multicolumn{1}{c|}{86.16} &
  \multicolumn{1}{c|}{78.62} &
  \multicolumn{1}{c|}{3} \\ \hline
\multicolumn{1}{|c|}{SecureID} &
  \multicolumn{1}{c|}{50.63} &
  \multicolumn{1}{c|}{90.94} &
  \multicolumn{1}{c|}{95.42} &
  \multicolumn{1}{c|}{99.42} &
  \multicolumn{1}{c|}{96.52} &
  \multicolumn{1}{c|}{6} \\ \hline
\multicolumn{7}{c}{\textbf{PAD-IDCard 2024 Baseline Algorithms}} \\ \hline
\multicolumn{1}{|c|}{Baseline1} &
  \multicolumn{1}{c|}{4.58} &
  \multicolumn{1}{c|}{1.84} &
  \multicolumn{1}{c|}{4.20} &
  \multicolumn{1}{c|}{14.96} &
  \multicolumn{1}{c|}{9.10} &
  \multicolumn{1}{c|}{-} \\ \hline
\multicolumn{1}{|c|}{Baseline2} &
  \multicolumn{1}{c|}{7.17} &
  \multicolumn{1}{c|}{5.26} &
  \multicolumn{1}{c|}{9.78} &
  \multicolumn{1}{c|}{24.40} &
  \multicolumn{1}{c|}{16.18} &
  \multicolumn{1}{c|}{-} \\ \hline
\multicolumn{1}{|c|}{Baseline3} &
  \multicolumn{1}{c|}{9.02} &
  \multicolumn{1}{c|}{8.14} &
  \multicolumn{1}{c|}{13.28} &
  \multicolumn{1}{c|}{28.58} &
  \multicolumn{1}{c|}{19.90} &
  \multicolumn{1}{c|}{-} \\ \hline
\end{tabular}%
}
\end{table}

As we mentioned before, the sequestered dataset was also analysed by country, separated by Chile, Guatemala, Panama, and Mexico. 

\begin{itemize}

\item  For the Chilean ID card, the IDVC\_V2 team achieved the best results by far, with a lower Average rank of 4.34\%. The screen attack was identified as the most challenging PAI.

\item For the Guatemala ID card, the IDVC\_V1 team reached the best results with an average rank of 41.25\%. 

\item For the Panama ID card, the Anonymous team reached the best results, with an average rank of 37.58\%. 

\item For Mexico, the Anonymous team reached the best results with an average rank of 76.94\%.
\end{itemize}
\vspace{-0.3cm}

It is essential to highlight that all the submission models reached the highest error (average rank) compared with baseline 1 trained with the MobileVIT model.

Tables \ref{tab:chile}, \ref{tab:guatemala}, \ref{tab:panama}, and \ref{tab:mexico} show all the submissions evaluated in a single country as complementary information for Chile, Guatemala, Panama, and Mexico respectively. For each table, the best results are shown in bold. 

For the Asmodeus team, estimating the EER, BPCER\textsubscript{10}, BPCER\textsubscript{20}, and BPCER\textsubscript{100} values was not possible because the models always delivered the same score values (0.86 and 0.99). These scores were also checked using validation scores directly with the team.

\begin{table}[]
\centering
\caption{Summary results on \textbf{Chile}}
\label{tab:chile}
\resizebox{\columnwidth}{!}{%
\begin{tabular}{|c|c|c|c|c|c|}
\hline
\begin{tabular}[c]{@{}c@{}}Team\\ Name\end{tabular} &
  \begin{tabular}[c]{@{}c@{}}EER\\ (\%)\end{tabular} &
  \begin{tabular}[c]{@{}c@{}}BPCER10\\ (\%)\end{tabular} &
  \begin{tabular}[c]{@{}c@{}}BPCER20\\ (\%)\end{tabular} &
  \begin{tabular}[c]{@{}c@{}}BPCER100\\ (\%)\end{tabular} &
  \begin{tabular}[c]{@{}c@{}}Average\\ Rank\\ (\%)\end{tabular} \\ \hline
Anonymous\_V1     & 16.00 & 29.80 & 47.00 & 79.30 & 59.71 \\ \hline
Anonymous\_V2     & 16.31 & 28.10 & 44.50 & 67.60 & 52.77 \\ \hline
Asmodeus\_V1        & N/A & N/A & N/A & N/A & N/A  \\ \hline
Asmodeus\_V2        & N/A & N/A & N/A & N/A & N/A \\ \hline
FRIFE               & 38.13 & 83.80 & 93.60 & 100 & 94.84 \\ \hline
IDVC\_V1 & 4.90 & 1.30 & 4.70 & 14.80 & 9.07 \\ \hline
IDVC\_V2 & 3.00 & 0.60 & 1.90 & 7.30 & \textbf{4.34} \\ \hline
Secure-ID           & 42.46 & 83.10 & 90.70 & 98.80 & 93.23  \\ \hline
Baseline1           & 0.70 & 0.01 & 0.01 & 0.50 & 0.25 \\ \hline
Baseline2           & 3.10 & 0.70 & 1.90 & 6.00 & 3.71 \\ \hline
Baseline3           & 3.98 & 1.50 & 3.20 & 8.70 & 5.61 \\ \hline
\end{tabular}%
}
\end{table}

\begin{table}[]
\centering
\caption{Summary results on \textbf{Guatemala}}
\label{tab:guatemala}
\resizebox{\columnwidth}{!}{%
\begin{tabular}{|c|c|c|c|c|c|}
\hline
\begin{tabular}[c]{@{}c@{}}Team\\ Name\end{tabular} &
  \begin{tabular}[c]{@{}c@{}}EER\\ (\%)\end{tabular} &
  \begin{tabular}[c]{@{}c@{}}BPCER10\\ (\%)\end{tabular} &
  \begin{tabular}[c]{@{}c@{}}BPCER20\\ (\%)\end{tabular} &
  \begin{tabular}[c]{@{}c@{}}BPCER100\\ (\%)\end{tabular} &
  \begin{tabular}[c]{@{}c@{}}Average\\Rank \\ (\%)\end{tabular} \\ \hline
Anonymous\_V1     & 15.30 & 22.10 & 33.90 & 62.80 & 45.99 \\ \hline
Anonymous\_V2     & 24.88 & 54.50 & 69.01 & 88.90 & 73.05 \\ \hline
Asmodeus\_V1        & N/A & N/A & N/A & N/A & N/A \\ \hline
Asmodeus\_V2        & N/A & N/A & N/A & N/A & N/A \\ \hline
FRIFE               & 47.88 & 93.30 & 97.30 & 100 & 97.85 \\ \hline
IDVC\_V1 & 6.38 & 2.10 & 9.60 & 75.90 & \textbf{41.25} \\ \hline
IDVC\_V2 & 15.88 & 22.30 & 34.00 & 77.30 & 53.31 \\ \hline
SecureID           & 50.75 & 92.00 & 95.40 & 99.20 &  96.62 \\ \hline
Baseline1           & 2.50 & 0.30 & 1.00 & 5.10 & 2.91 \\ \hline
Baseline2           & 7.90 & 6.80 & 10.10 & 20.20 & 14.49 \\ \hline
Baseline3           & 13.69 & 15.40 & 21.50 & 32.90 & 25.98 \\ \hline
\end{tabular}%
}
\end{table}

\begin{table}[]
\centering
\caption{Summary results on \textbf{Panama}}
\label{tab:panama}
\resizebox{\columnwidth}{!}{%
\begin{tabular}{|c|c|c|c|c|c|}
\hline
\begin{tabular}[c]{@{}c@{}}Team\\ Name\end{tabular} &
  \begin{tabular}[c]{@{}c@{}}EER\\ (\%)\end{tabular} &
  \begin{tabular}[c]{@{}c@{}}BPCER10\\ (\%)\end{tabular} &
  \begin{tabular}[c]{@{}c@{}}BPCER20\\ (\%)\end{tabular} &
  \begin{tabular}[c]{@{}c@{}}BPCER100\\ (\%)\end{tabular} &
  \begin{tabular}[c]{@{}c@{}}Average\\ Rank\\ (\%)\end{tabular} \\ \hline
Anonymous\_V1     & 13.06 & 16.20 & 27.80 & 52.00 & \textbf{37.58} \\ \hline
Anonymous\_V2     & 15.90 & 27.20 & 47.80 & 85.50 & 62.53 \\ \hline
Asmodeus\_V1        & N/A & N/A & N/A & N/A & N/A \\ \hline
Asmodeus\_V2        & N/A & N/A & N/A & N/A & N/A \\ \hline
FRIFE               & 41.68 & 92.00 & 94.60 & 99.99 & 96.77 \\ \hline
IDVC\_V1 & 10.78 & 13.40 & 44.50 & 86.20 & 59.13 \\ \hline
IDVC\_V2 & 18.78 & 30.60 & 40.50 & 69.00 & 52.77 \\ \hline
SecureID           & 53.65 & 93.90 & 96.90 & 99.60 & 97.47  \\ \hline
Baseline1           & 7.20 & 6.40 & 10.40 & 20.10 & 14.45 \\ \hline
Baseline2           & 12.98 & 15.80 & 22.70 & 32.80 & 26.37 \\ \hline
Baseline3           & 11.78 & 13.10 & 18.00 & 27.50 & 21.77 \\ \hline
\end{tabular}%
}
\end{table}

\begin{table}[]
\centering
\caption{Summary results on \textbf{Mexico}}
\label{tab:mexico}
\resizebox{\columnwidth}{!}{%
\begin{tabular}{|c|c|c|c|c|c|}
\hline
\begin{tabular}[c]{@{}c@{}}Team\\ Name\end{tabular} &
  \begin{tabular}[c]{@{}c@{}}EER\\ (\%)\end{tabular} &
  \begin{tabular}[c]{@{}c@{}}BPCER10\\ (\%)\end{tabular} &
  \begin{tabular}[c]{@{}c@{}}BPCER20\\ (\%)\end{tabular} &
  \begin{tabular}[c]{@{}c@{}}BPCER100\\ (\%)\end{tabular} &
  \begin{tabular}[c]{@{}c@{}}Average\\ Rank\\ (\%)\end{tabular} \\ \hline
Anonymous\_V1     & 22.48 & 51.80 & 71.10 & 90.50 & \textbf{76.94} \\ \hline
Anonymous\_V2     & 29.88 & 64.30 & 72.90 & 88.20 & 78.83 \\ \hline
Asmodeus\_V1        & N/A & N/A & N/A & N/A & N/A \\ \hline
Asmodeus\_V2        & N/A & N/A & N/A & N/A & N/A \\ \hline
FRIFE               & 46.86 & 86.40 & 94.40 & 99.40 & 95.3 \\ \hline
IDVC\_V1 & 24.70 & 65.90 & 83.30 & 97.80 & 87.07 \\ \hline
IDVC\_V2 & 30.28 & 73.40 & 87.10 & 96.50 & 89.06 \\ \hline
SecureID           & 57.08 & 96.10 & 98.40 & 99.90 &  98.69 \\ \hline
Baseline1           & 5.80 & 2.20 & 6.70 & 27.30 & 16.10 \\ \hline
Baseline2           & 2.38 & 0.50 & 0.70 & 5.70 & 3.16 \\ \hline
Baseline3           & 4.10 & 0.80 & 3.00 & 20.50 & 11.31 \\ \hline
\end{tabular}%
}
\end{table}

\newpage

\vspace{-0.3cm}

\section{Analysis}
The competition results show that the generalisation capabilities to predict PAD between different countries and attacks are still challenging. According to our analysis, the number of images available for training in open datasets limits the performance of the approaches. The open-set datasets present fewer bona fide images per subject or use a printed PVC ID card to simulate a genuine image. Further on, the same ID card is used to create many attacks. As a result, imbalanced datasets are obtained. This factor confuses the classifier with the print and screen attack, obtaining a high EER. Conversely, the teams that used private datasets based on ID cards with many different subjects and reduced the number of images per subject in the datasets obtained the best results. The screen attack (blue curve) is identified as the most challenging in the baseline and for the team which achieved the best result. The high resolution of different screens available in the market makes this attack very hard to detect.

The different security factors present on ID cards, such as holograms, watermarks, and others, were created based on a physical inspection using active factors and lights. Thus, these factors are not a challenge that is easily reproducible in a PAD system based on one image in remote systems.
 
We can also identify that ID cards in countries based on ICAO compliance standards are more accessible to detect and classify, such as Chile and Panama. The standardised position of the face photo, letter sizes, and other factors support the learning process. Conversely, Guatemala and Mexico ID cards do not follow the ICAO standards and present a lot of variability in photos, illumination, and where the different information is positioned. A general and agnostic system is still a challenge, as demonstrated in the test evaluation results for specific countries. 

\section{Conclusion}
\label{sec:conclu}


The results from this competition indicate that ID card PAD is still far away from fully solving this research challenge. Significant differences in accuracy among baseline algorithms, which were trained with different data considering private and open-set datasets, stress the importance of access to extensive and diversified training datasets encompassing a large number of PAIs. 

As a future work, we will propose a new version of the competition based on synthetic ID cards to reduce the lack of bona fide images and attacks. In order to measure the generalisation capabilities, new approaches based on meta-learning approaches, such as zero-shot and few-shot learning, are suggested to improve the generalisation capabilities of countries not included in the training set.

This competition and the benchmark will contribute to our efforts as a biometric community to win the PAD arms race.

\section*{Acknowledgements}
This competition was sponsored by Facephi, R\&D area. Further, this work was supported by the European Union’s Horizon 2020 research and innovation program under grant agreements 883356 (iMARS) and 101121280 (EINSTEIN), and the German Federal Ministry of Education and Research and the Hessian Ministry of Higher Education, Research, Science and the Arts within their joint support of the National Research Center for Applied Cybersecurity ATHENE.


{\small
\bibliographystyle{ieee}
\bibliography{egbib}
}

\end{document}